\documentclass{article} 
\usepackage{iclr2022_conference,times}

\usepackage{amsmath,amsfonts,bm}

\def\eqref#1{equation~\ref{#1}}

\def\1{\bm{1}}

\DeclareMathAlphabet{\mathsfit}{\encodingdefault}{\sfdefault}{m}{sl}
\SetMathAlphabet{\mathsfit}{bold}{\encodingdefault}{\sfdefault}{bx}{n}

\usepackage{multirow}
\usepackage{hyperref}
\usepackage{url}
\usepackage{booktabs}
\usepackage{enumitem}
\usepackage{mathtools}
\usepackage{bbm}
\usepackage{xspace}
\usepackage{authblk}

\usepackage[normalem]{ulem}
\useunder{\uline}{\ul}{}

\newcommand*{\affaddr}[1]{#1} 
\newcommand*{\affmark}[1][*]{\textsuperscript{#1}}
\newcommand*{\email}[1]{\texttt{#1}}

\title{Improving Contrastive Learning with Model Augmentation}

\author{\textbf{Zhiwei~Liu}\affmark[1], \textbf{Yongjun~Chen}\affmark[1], \textbf{Jia~Li}\affmark[1], \textbf{Man~Luo}\affmark[2], \textbf{Philip S. ~Yu}\affmark[3], and \textbf{Caiming~Xiong}\affmark[1]\\
\affaddr{\affmark[1]Salesforce Research, CA, USA} \\ \email{\{zhiweiliu, yongjun.chen, jia.li, cxiong\}@salesforce.com}\\
\affaddr{\affmark[2]Arizona State University, AZ, USA; \email{mluo26@asu.edu}} \\
\affaddr{\affmark[3]University of Illinois at Chicago, IL, USA; \email{psyu@uic.edu}}
}

\newcommand{\modelname}{\textsc{SRMA}\xspace}

\iclrfinalcopy 
\begin{document}

\maketitle

\begin{abstract}
The sequential recommendation aims at predicting the next items in user behaviors, which can be solved by characterizing item relationships in sequences. Due to the data sparsity and noise issues in sequences, a new self-supervised learning (SSL) paradigm is proposed to improve the performance, which employs contrastive learning between positive and negative views of sequences. 
However, existing methods all construct views by adopting augmentation from data perspectives, while we argue that 1) optimal data augmentation methods are hard to devise, 2) data augmentation methods destroy sequential correlations, and 3) data augmentation fails to incorporate comprehensive self-supervised signals. 
Therefore, we investigate the possibility of model augmentation to construct view pairs. We propose three levels of model augmentation methods: neuron masking, layer dropping, and encoder complementing. 
This work opens up a novel direction in constructing views for contrastive SSL. Experiments verify the efficacy of model augmentation for the SSL in the sequential recommendation. Code is available\footnote{\url{https://github.com/salesforce/SRMA}}.
\end{abstract}

\section{Introduction}
The sequential recommendation~\citep{fan2021continuous,liu2021augmenting,chen2018sequential,tang2018personalized,zheng2019gated} aims at predicting future items in sequences, where the crucial part is to characterize item relationships in sequences. Recent developments in sequence modeling~\citep{fan2021continuous,liu2021augmenting} verify the superiority of Transform~\citep{vaswani2017attention}, i.e. the self-attention mechanism, in revealing item correlations in sequences.  
A Transformer~\citep{kang2018self} is able to infer the sequence embedding at specified positions by weighted aggregation of item embeddings, where the weights are learned via self-attention. 
Existing works~\citep{fan2021continuous,ssept20wu} further improve Transformer by incorporating additional complex signals. 


However, the \text{data sparsity issue}~\citep{liu2021augmenting} and \text{noise in sequences} undermine the performance of a model in sequential recommendation. 
The former hinders performance due to insufficient training since the complex structure of a sequential model requires a dense corpus to be adequately trained. 
The latter also impedes the recommendation ability of a model because noisy item sequences are unable to reveal actual item correlations. 
To overcome both, a new contrastive self-supervised learning (SSL) paradigm~\citep{liu2021contrastive,xie2020contrastive,zhou2020s3} is proposed recently.
This paradigm enhances the capacity of encoders by leveraging additional self-supervised signals. Specifically,
the SSL paradigm constructs positive view pairs as two data augmentations from the same sequences~\citep{xie2020contrastive}, while negative pairs are augmentations from distinct sequences. 
Incorporating augmentations during training increases the amount of training data, thus alleviating the sparsity issue. And the contrastive loss~\citep{chen2020simple} improves the robustness of the model, which endows a model with the ability to against noise.  

Though being effective in enhancing sequential modeling, 
the data augmentation methods adopted in the existing SSL paradigm suffer from the following weaknesses:
\begin{itemize}[leftmargin=*]
    \item Optimal data augmentation methods are hard to devise. Current sequence augmentation methods adopts random sequence perturbations~\citep{liu2021contrastive,xie2020contrastive}, 
    which includes \text{crop}, \text{mask}, \text{reorder}, \text{substitute} and \text{insert} operations. 
    Though a random combination of those augmenting operations improves the performance, it is rather time-consuming to search the optimal augmentation methods from a large number of potential combinations for different datasets~\citep{liu2021contrastive}.
    \item Data augmentation methods destroy sequential correlations, leading to less confident positive pairs. 
    The existing SSL paradigm requires injecting perturbations into the augmented views of sequences for contrastive learning. 
    However, because the view construction process is not optimized to characterize sequential correlations,
    two views of one sequence may reveal distinct item relationships, which should not be recognized as positive pairs.  
    \item Data augmentation fails to incorporate comprehensive self-supervised signals. Current data augmentation methods are designed based on heuristics, which already requires additional prior knowledge. Moreover, since the view construction process is not optimized with the encoder, data augmentation may only reveal partial self-supervised signals from data perspectives. Hence, we should consider other types of views besides data augmentation. 

\end{itemize}

Therefore, we investigate the possibility of \textit{model augmentation} to construct view pairs for contrastive learning, which functions as a complement to the data augmentation methods. 
We hypothesis that injecting perturbations into the encoder should enhance the self-supervised learning ability to existing paradigms. 
The reasons are threefold: Firstly, model augmentation is jointly trained with the optimization process, thus endows the end-to-end training fashion. As such, it is easy to discover the optimal view pairs for contrastive learning. 
Secondly, model augmentation constructs views without manipulation to the original data, which leads to high confidence of positive pairs. 
Last but not least, injecting perturbation into the encoder has distinct characteristics to data augmentation, which should be an important complement in constructing view pairs for existing self-supervised learning scheme~\citep{liu2021contrastive,zhou2020s3}. 

This work studies the model augmentation for a self-supervised sequential recommendation from three levels: 1) neuron masking (dropout), which adopts the dropout layer to randomly mask partial neurons in a layer. By operating the dropout twice to one sequence, we can perturb the output of the embedding from this layer, which thus constructs two views from model augmentation perspective~\citep{gao2021simcse}.  2) layer dropping. Compared with neuron masks, we randomly drop a complete layer in the encoder to inject more perturbations. By randomly dropping layers in an encoder twice, we construct two distinct views. Intuitively,  layer-drop augmentation enforces the contrast between deep features and shallows features of the encoder. 
3) encoder complementing, which leverages other encoders to generate sequence embeddings. Encoder complementing augmentation is able to fuse distinct sequential correlations revealed by different types of encoders. For example, RNN-based sequence encoder~\citep{hidasi2015session} can better characterize direct item transition relationships, while Transformer-based sequence encoder models position-wise sequential correlations. 
Though only investigating SSL for a sequential recommendation, we remark that model augmentation methods can also be applied in other SSL scenarios. The contributions are as follows:
\begin{itemize}[leftmargin=*]
    \item We propose a new contrastive SSL paradigm for sequential recommendation by constructing views from model augmentation, which is named as \modelname. 
    \item We introduce three levels of model augmentation methods for constructing view pairs. 
    \item We discuss the effectiveness and conduct a comprehensive study of model augmentations for the sequential recommendation. 
    \item We investigate the efficacy of different variants of model augmentation. 
    
\end{itemize}
 
\section{Related Work}
\subsection{Sequential Recommendation}
Sequential recommendation predicts future items in user 
sequences by encoding sequences while modeling item transition correlations~\citep{rendle2010factorizing,hidasi2015session}. Previously, Recurrent Neural Network~(RNN) have been adapted to sequential recommendation~\citep{hidasi2015session,wu2017recurrent}, 
ostensibly modeling sequence-level item transitions.
Hierarchical RNNs~\citep{quadrana2017personalizing} incorporate
personalization
information. 
Moreover, both 
long-term and short-term item transition correlations are modelled in LSTM~\citep{wu2017recurrent} . 
Recently, the success of self-attention models~\citep{vaswani2017attention,devlin2018bert} promotes the prosperity of Transformer-based sequential recommendation models. SASRec~\citep{kang2018self} is a pioneering work adapting Transformer to 
characterize complex item transition correlations.
BERT4Rec~\citep{sun2019bert4rec} adopts the bidirectional Transformer layer to encode sequence. 
ASReP~\citep{liu2021augmenting} reversely pre-training a Transformer to augment short sequences and fine-tune it to predict the next-item in sequences. TGSRec~\citep{fan2021continuous} models temporal collaborative signals in sequences to recognize item relationships. 

\subsection{Self-supervised Learning}
Self-supervised learning~(SSL) is proposed recently to describe ``the machine predicts any parts of its input for
any observed part''\citep{Bengio2021deep}, which stays within the narrow scope of unsupervised learning. 
To achieve the self-prediction, endeavors from various domains have developed different SSL schemes from either generative or contrastive perspectives~\citep{liu2021self}. For generative SSL, the masked language model is adopted in BERT~\citep{devlin2018bert} to generate masked words in sentences. GPT-GNN~\citep{hu2020gpt} also generates masked edges to realize SSL. Other generative SSL paradigms in computer vision~\citep{oord2016conditional} are proposed.
Compared with generative SSL, contrastive SSL schemes have demonstrated more promising performance. 
SimCLR~\citep{chen2020simple} proposes simple contrastive learning between augmented views for images, which is rather effective in achieving SSL. 
GCC~\citep{qiu2020gcc} and GraphCL~\citep{you2020graph} adopts contrastive learning between views from corrupted graph structures. 
CL4SRec~\citep{xie2020contrastive} and CoSeRec~\citep{liu2021contrastive} devise the sequence augmentation methods for SSL on sequential recommendation. 
This paper also investigates the contrastive SSL for a sequential recommendation. Instead of adopting the data augmentation for constructing views to contrast, we propose the model augmentation to generate contrastive views.

\section{Preliminary}
\subsection{Problem Formulation}
We denote 
user and item sets as $\mathcal{U}$ and $\mathcal{V}$ respectively. Each user $u\in \mathcal{U}$ is associated with a sequence of items in chronological order
$s_{u}= [v_{1}, \dots, v_{t}, \dots, v_{|s_{u}|}]$, where $v_{t}\in \mathcal{V}$ denotes the item that $u$
has interacted with at time $t$ and $|s_{u}|$ is the total number of items. 
Sequential recommendation is formulated as follows:
\begin{equation}
\underset{v_{i}\in \mathcal{V}}{\mathrm{arg\,max}}~P(v_{|s_{u}|+1}=v_{i}\left| s_{u}\right.),
\end{equation}
where $v_{|s_{u}|+1}$ denotes
the next item in sequence.
Intuitively, we calculate the probability of all candidate items and recommend items with high probability scores.

\subsection{Sequential Recommendation Framework}
The core of a generic sequential recommendation framework is a sequence encoder $\mathsf{SeqEnc}(\cdot)$, which transforms item sequences to embeddings for scoring. We formulate the encoding step as: 
\begin{equation}\label{eq:seq_encoder}
\mathbf{h}_{u} = \mathsf{SeqEnc}(s_{u}),
\end{equation}
where $\mathbf{h}_{u}$ denotes the sequence embedding of $s_{u}$. To be specific, if we adopt a Transformer~\citep{kang2018self,vaswani2017attention} as the encoder, $\mathbf{h}_{u}$ is a bag of embeddings, where
at each position $t$,
$\mathbf{h}_{u}^{t}$, represents a predicted next-item.
We adopt the log-likelihood loss function to optimize the encoder for next-item prediction as follows:
\begin{equation}
\mathcal{L}_{\text{rec}}(u,t) = -\log(\sigma(\mathbf{h}_{u}^{t}\cdot \mathbf{e}_{v_{t+1}}))- \sum_{v_{j}\not\in s_{u}}\log(1-\sigma (\mathbf{h}_{u}^{t} \cdot \mathbf{e}_{v_{j}})),
\end{equation}
where $\mathcal{L}_{\text{rec}}(u,t)$ denotes the loss score for the prediction at position $t$ in sequence $s_{u}$, $\sigma$ is the non-linear activation function, $\mathbf{e}_{v_{t+1}}$ denotes the embedding for item $v_{t+1}$, and $v_{j}$ is the sampled negative item for $s_u$. 
The embeddings of items are retrieved from the embedding layer in $\mathsf{SeqEnc}(\cdot)$, which is jointly optimized with other layers. 

\subsection{Contrastive Self-supervised Learning Paradigm}
\begin{figure}
    \centering
    \includegraphics[width=0.99\linewidth]{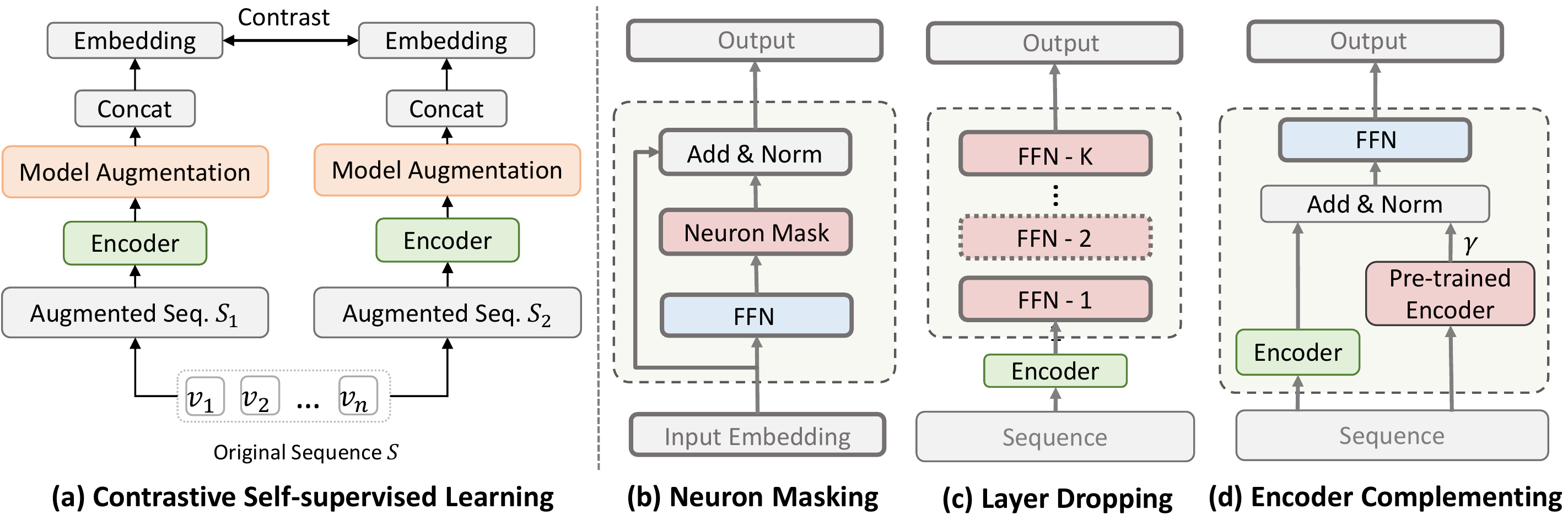}
    \caption{(a) The contrastive SSL framework with model augmentation. We apply the model augmentation to the encoder, which constructs two views for contrastive learning.  (b) the neuron masking augmentation. We demonstrate the neuron masking for the Feed-Forward network. (c) the layer dropping augmentation. We add $K$ FFN layers after the encoder and randomly drop $M$ layers (dash blocks) during each batch of training. And (d) the encoder complementing augmentation. We pre-train another encoder for generating the embedding of sequences. The embedding from the pre-trained encoder is combined with the model encoder for contrastive learning. 
    }
    \label{fig:framework}
\end{figure}

Other than the next-item prediction, we can leverage other pretext tasks~\citep{sun2019bert4rec,liu2021self,liu2021contrastive} over the sequence to optimize the encoder, which harnesses the self-supervised signals within the sequence.   
This paper investigate the widely adopted contrastive SSL scheme~\citep{liu2021contrastive,xie2020contrastive}. This scheme constructs positive and negative view pairs from sequences, and employs the contrastive loss~\citep{oord2018representation} to optimize the encoder. We formulate  the SSL step as follows:
\begin{equation}
\mathcal{L}_{\mathrm{ssl}}(\mathbf{\Tilde{h}}_{2u-1}, 
    \mathbf{\tilde{h}}_{2u}) = 
    - \log \frac{\exp(\text{sim}(\mathbf{\tilde{h}}_{2u-1},                       \mathbf{\tilde{h}}_{2u}))}
    {\sum_{m=1}^{2N}\mathbbm{1}_{m\neq 2u-1}\exp(\text{sim}(\mathbf{\tilde{h}}_{2u-1}, \mathbf{\tilde{h}}_{m}))},
\end{equation}  
where $\mathbf{\tilde{h}}_{2u}$ and $\mathbf{\tilde{h}}_{2u-1}$ denotes two views constructed for sequence $s_{u}$.  $\mathbbm{1}$ is an indication function. $\text{sim}(\cdot,\cdot)$ is the similarity function, \textit{e.g.} dot-product. Since each sequence has two view, we have $2N$ samples in a batch with $N$ sequences for training. The nominator indicates the agreement maximization between a positive pair, while the denominator is interpreted as push away those negative pairs. 
Existing works apply data augmentation for sequences to construct views, \textit{e.g.} \cite{xie2020contrastive} propose \textit{crop}, \textit{mask}, and \textit{reorder} a sequence and \citep{liu2021contrastive} devises \textit{insert} and \textit{substitute} sequence augmentations. For sequential recommendation, since both SSL and the next-item prediction charaterize the item relationships in sequences, we add them together to optimize the encoder. Therefore, the final loss $\mathcal{L} = \mathcal{L}_{\mathrm{rec}} + \lambda \mathcal{L}_{\mathrm{ssl}}.$ 
Compared with them, we adopt both the data augmentation and model augmentation to generate views for contrast.
We demonstrate the contrastive SSL step in Figure~\ref{fig:framework}(a).

\section{Model Augmentation}
In this section, we introduce the model augmentation to construct views for sequences. We discuss three type of model augmentation methods, which are neuron mask, layer drop and encoder complement. We illustrate these augmentation methods in Figure~\ref{fig:framework}(b), \ref{fig:framework}(c) and \ref{fig:framework}(d), respectively.

\subsection{Neuron Masking}
This work adopts the Transformer as the sequence encoder, which passes the hidden embeddings to the next layer through a feed-forward network (FFN). During training, we randomly mask partial neurons in each FFN layer, which involves a masking probability $p$. The large value of $p$ leads to intensive embedding perturbations. As such, we generate a pair of views from one sequence from model perspectives. 
Besides, during each batch of training, the masked neurons are randomly selected, which results in comprehensive contrastive learning on model augmentation.
Note that, though we can utilize different probability values for distinct FFN layers, we enforce their neuron masking probability to be the same for simplicity. 
The neuron masking augmentation on FFN is shown in Figure~\ref{fig:framework}(b).  
Additionally, we remark that the neuron mask can be applied to any neural layers in a model to inject more perturbations.  

\subsection{Layer Dropping}
Dropping partial layers of a model decreases the depth and reduces complexity. Previous research argues that most recommender systems require only shallow embeddings for users and items~\citep{dacrema2019we}. 
Therefore, it is reasonable to randomly drop a fraction of layers during training, which functions as a way of regularization.
Additionally, existing works~\citep{liu2020towards,he2016deep} claim that embeddings at shallow layers and deep layers are both important to reflect the comprehensive information of the data. 
Dropping layers enable contrastive learning between shallow embeddings and deep embeddings, thus being an enhancement of existing works that only contrasting between deep features. 

On the other hand, dropping layers, especially those necessary layers in a model, may destroy original sequential correlations. Thus, views generating by dropping layers may not be a positive pair. 
To this end, instead of manipulating the original encoder, we stack $K$  FFN layers after the encoder and randomly drop $M$ of them during each batch of training, where $M < K$. 
We illustrate the layer dropping as in Figure~\ref{fig:framework}(c), where we append $K$ additional FFN layers after the encoder and use dash blocks to denote the dropped layers. 

\subsection{Encoder Complementing}
During self-supervised learning, we employ one encoder to generate embeddings of two views of one sequence. 
Though this encoder can be effective in revealing complex sequential correlations, contrasting on one single encoder may result in embedding collapse problems for self-supervised learning~\citep{hua2021feature}. 
Moreover, one single encoder is only able to reflect the item relationships from a unitary perspective. For example, the Transformer encoder adopts the attentive aggregation of item embeddings to infer sequence embedding, while an RNN structure~\citep{hidasi2015session} is more suitable in encoding direct item transitions. Therefore,
contrasting between views from distinct encoders enables the model to learn comprehensive sequential relationships of items.

However, embeddings from two views of a sequence with distinct encoders lead to a non-Siamese paradigm for self-supervised learning, which is hard to train and suffers the embedding collapse problem~\citep{koch2015siamese,chen2021exploring}. Additionally, if two distinct encoders reveal significantly diverse sequential correlations, the embeddings are far away from each other, thus being bad views for contrastive learning~\citep{tian2020makes}. Moreover, though we can optimize two encoders during a training phase, it is still problematic to combine them for the inference of sequence embeddings to conduct recommendations. 

As a result, instead of contrastive learning with distinct encoders, we harness another pre-trained encoder as an encoder complementing model augmentation for the original encoder. To be more specific, we first pre-train another encoder with the next-item prediction target. Then, in the self-supervised training stage, we utilize this pre-trained encoder to generate another embedding for a view. After that, we add the view embeddings from a model encoder and the pre-trained encoder. 
We illustrate the encoder complementing augmentation in Figure~\ref{fig:framework}(d).
Note that we only apply this model augmentation in one branch of the SSL paradigm. And the embedding from the pre-trained encoder is re-scaled by a hyper-parameter $\gamma$ before adding to the embedding from the framework's encoder. The smaller value of $\gamma$ implies injecting fewer perturbations from a distinct encoder. 
The pre-trained encoder are not trainable during training.
Hence, there is no optimization for this pre-trained encoder and it is no longer required to take account of both encoders during the inference stage. 

\section{Experiments}

\subsection{Experimental Settings}

\textbf{Dataset}~ We conduct experiments on three public datasets. Amazon Sports, Amazon Toys and Games~\citep{mcauley2015image} and Yelp\footnote{https://www.yelp.com/dataset}, which
are Amazon review data in Sport and Toys categories, and
a dataset for the business recommendation, respectively.
We follow common 
practice
in~\citep{liu2021augmenting,xie2020contrastive} to only keep the `5-core' sequences.  
In total, Sports dataset has 35,598 users, 18,357 items and 296,337 interactions. Toys dataset contains 19,412 users, 11,924 items, and 167,597interactions. Yelp dataset consists 30,431 users, 20,033 items and 316,354 interactions. 

\textbf{Evaluation Metrics}~
We follow existing works~\citep{wang2019neural,krichene2020sampled,liu2021augmenting} 
to evaluate models' performances based on 
the whole item set without negative sampling 
and report standard 
\textit{Hit Ratio}$@k$ ($\mathrm{HR}@k$) and \textit{Normalized Discounted 
Cumulative Gain}$@k$ ($\mathrm{NDCG}@k$) 
on 
all datasets, where $k\in\{5, 10, 20\}$.

\textbf{Baselines}~
We include two groups of sequential models as
baselines 
for comprehensive comparisons.
The first group baselines are sequential models that
use different
deep neural architectures to encode sequences with
a supervised objective. These include
\textbf{GRU4Rec}~\citep{hidasi2015session} as an RNN-based method, 
\textbf{Caser}~\citep{tang2018personalized} as a
CNN-based approach, and \textbf{SASRec}~\citep{kang2018self} 
as one of the state-of-the-art Transformer based
solution.
The second group baselines additionally
leverage SSL objective.
\textbf{BERT4Rec}~\citep{sun2019bert4rec} employs
a \emph{Cloze} task~\citep{taylor1953cloze} as a generative self-supervised learning sigal.
\textbf{S$^3\text{Rec}$}~\citep{zhou2020s3} 
uses contrastive SSL with `mask' data augmentation to fuse correlation-ships
among item, sub-sequence, and correspondinng
attributes into the networks.
We remove the components for
fusing attributes for fair comparison.
\textbf{CL4SRec}~\citep{xie2020contrastive} maximize the agreements
between two sequences augmentation, where the data augmentation are randomly selected from  
'crop', `reorder', and `mask' data augmentations. \textbf{CoSeRec}~\citep{liu2021contrastive}
improves the robustness of
data augmentation under
contrastive SSL framework by leveraging
item-correlations.

\textbf{Implementation Details}
The model encoder in \modelname is the basic Transformer-based encoder. We adopt the widely used SASRec encoder. The neuron masking probability is searched from $\{0.0,0.1,0.2,\dots,0.9\}$. For layer dropping, the $K$ is searched from $\{1,2,3,4\}$, and $M$ is searched accordingly. As for encoder complementing, we search the re-scale hyper-parameter $\gamma$
from $\{0.005, 0.01, 0.05, 0.1, 0.5, 1.0\}$ and the pre-trained encoder is selected from a $1$-layer Transformer and a GRU encoder.

\subsection{Overall Performance}
\begin{table*}
  \caption{Performance comparisons of different methods. 
  The best score is in bold in each row, and the second best is underlined.}
  \vspace{2mm}
\label{tab:main-results}
\setlength{\tabcolsep}{0.65mm}{
\begin{tabular}{l|l|cccccccr}
\toprule
Dataset                 & Metric  & GRU4Rec & Caser  & SASRec & BERT4Rec & S$^3$-Rec & CL4SRec & CoSeRec         & \modelname            \\
\hline
\hline
\multirow{6}{*}{Sports} & HR@5    & 0.0162  & 0.0154 & 0.0206 & 0.0217   & 0.0121                   & 0.0231  & {\ul 0.0287}    & \textbf{0.0299} \\
& HR@10   & 0.0258  & 0.0261 & 0.0320  & 0.0359   & 0.0205                   & 0.0369  & {\ul 0.0437}    & \textbf{0.0447} \\
& HR@20   & 0.0421  & 0.0399 & 0.0497 & 0.0604   & 0.0344                   & 0.0557  & {\ul 0.0635}    & \textbf{0.0649} \\
& NDCG@5  & 0.0103  & 0.0114 & 0.0135 & 0.0143   & 0.0084                   & 0.0146  & {\ul 0.0196}    & \textbf{0.0199} \\
& NDCG@10 & 0.0142  & 0.0135 & 0.0172 & 0.019    & 0.0111                   & 0.0191  & {\ul 0.0242}    & \textbf{0.0246} \\
& NDCG@20 & 0.0186  & 0.0178 & 0.0216 & 0.0251   & 0.0146                   & 0.0238  & {\ul 0.0292}    & \textbf{0.0297} \\
\midrule
\multirow{6}{*}{Yelp}   & HR@5    & 0.0152  & 0.0142 & 0.0160  & 0.0196   & 0.0101                   & 0.0227  & {\ul 0.0241}    & \textbf{0.0243} \\
& HR@10   & 0.0248  & 0.0254 & 0.0260  & 0.0339   & 0.0176                   & 0.0384  & {\ul 0.0395}    & \textbf{0.0395} \\
& HR@20   & 0.0371  & 0.0406 & 0.0443 & 0.0564   & 0.0314                   & 0.0623  & \textbf{0.0649} & {\ul 0.0646}    \\
& NDCG@5  & 0.0091  & 0.008  & 0.0101 & 0.0121   & 0.0068                   & 0.0143  & {\ul 0.0151}    & \textbf{0.0154} \\
& NDCG@10 & 0.0124  & 0.0113 & 0.0133 & 0.0167   & 0.0092                   & 0.0194  & {\ul 0.0205}    & \textbf{0.0207} \\
& NDCG@20 & 0.0145  & 0.0156 & 0.0179 & 0.0223   & 0.0127                   & 0.0254  & {\ul 0.0263}    & \textbf{0.0266} \\
\midrule
\multirow{6}{*}{Toys}   & HR@5    &  0.0097  & 0.0166 &  0.0463  & 0.0274 &  0.0143 & 0.0525 & {\ul 0.0583} & \textbf{0.0598} \\
& HR@10   & 0.0176  & 0.0270 & 0.0675  & 0.0450   & 0.0094  & 0.0776 & {\ul 0.0812} & \textbf{0.0834} \\
& HR@20   & 0.0301  & 0.0420 & 0.0941  & 0.0688   & 0.0235 &  0.1084  & {\ul 0.1103} & \textbf{0.1132}  \\
& NDCG@5 & 0.0059 & 0.0107 & 0.0306 & 0.0174 & 0.0123 & 0.0346 & {\ul 0.0399}  & \textbf{0.0407} \\
& NDCG@10 & 0.0084 & 0.0141 & 0.0374 & 0.0231 & 0.0391 & 0.0428
& {\ul 0.0473} & \textbf{0.0484} \\
& NDCG@20 & 0.0116 & 0.0179 & 0.0441 & 0.0291 & 0.0162 & 0.0505
&  {\ul 0.0547}  & \textbf{0.0559} \\

\bottomrule
\end{tabular}
}
\end{table*}

We compare the proposed paradigm \modelname to existing methods with respect to the performance on the sequential recommendation. Results are demonstrated in Table~\ref{tab:main-results}. We can observe that Transformer-based sequence encoders, such SASRec and BERT4Rec are better than GRU4Rec or Caser sequence encoders. Because of this, our proposed model \modelname also adopts the Transformer as sequence encoder. Moreover, the SSL paradigm can significantly improve performance. For example, the CL4SRec model, which adopts the random data augmentation, improves the performance of SASRec on HR and NDCG by $13.2\%$ and $9.8\%$ on average regarding the Sports dataset, respectively. 
Also, since \modelname enhances the SSL with both data augmentation and model augmentation, \modelname thus outperforms all other SSL sequential recommendation models. \modelname adopts the same data augmentation methods as CL4SRec. Nevertheless, \modelname significantly outperforms CL4SRec. On the sports dataset, we achieve $18.9\%$ and $27.9\%$ relative improvements on HR and NDCG, respectively. On the Yelp dataset, we achieve $4.5\%$ and $6.4\%$ relative improvements on HR and NDCG, respectively. And on Toys data, we achieve $8.6\%$ and $13.8\%$ relative improvements on HR and NDCG, respectively.
In addition, \modelname also performs better than CoSeRec which leverages item correlations for data augmentation. Those results all verify the effectiveness of model augmentation in improving the SSL paradigm. 

\subsection{Comparison between Model and Data Augmentation }
Because \modelname adopts the random sequence augmentation, we mainly focus on comparing with CL4SRec to justify the impacts of model augmentation and data augmentation. In fact, CL4SRec also implicitly uses the neuron masking model augmentation, where dropout layers are stacked within its original sequence encoder.
To separate the joint effects of model and data augmentation, we create its variants `CL4S. $p=0$', which sets all the dropout ratios to be $0$, thus disables the neuron masking augmentation. 
Also, another variant `CL4S. w/o D', which has no data augmentation are also compared. 
Additionally, we create two other variants of \modelname as `\modelname w/o M' and `\modelname w/o D' by disabling the model augmentation and data augmentation respectively. `\modelname w/o M' has additional FFN layers compared with `CL4S. $p=0$'. 
The recommendation performance on the Sports and Toys dataset is presented in Table~\ref{tab:variants-study}. 
We have the following observations. 
Firstly, we notice a significant performance drop of the variant `CL4S. $p=0$', which suggests that the neuron masking augmentation is rather crucial. It benefits both the regularization of the training encoder and model augmentation of SSL. 
Secondly, `\modelname w/o D' outperforms other baselines on the Sports dataset and has comparable performance to `CL4S.', which indicates the model augmentation is of more impact in the SSL paradigm compared with data augmentation.
Thirdly, \modelname performs the best against all the variants. This result suggests that we should jointly employ the data augmentation and model augmentation in an SSL paradigm, which contributes to comprehensive contrastive self-supervised signals.

\begin{table*}
\caption{Performance comparison w.r.t. the variants of CL4SRec (CL4S.) and \modelname. M and D denote the model augmentation and data augmentation, respectively. $p=0$ indicates no neuron masking. The best score in each column are in bold, where the second-best are underlined.}
\vspace{2mm}
\label{tab:variants-study}
\resizebox{\columnwidth}{!}{
\begin{tabular}{l|cc|cc|cc|cc}
\toprule
\multirow{3}{*}{Model} & \multicolumn{4}{c|}{Sports} 
 & \multicolumn{4}{c}{Toys}\\
\cline{2-9}
& \multicolumn{2}{c|}{HR} &
\multicolumn{2}{c|}{NDCG} &
\multicolumn{2}{c|}{HR} &
\multicolumn{2}{c}{NDCG} \\ 
& @5 & @10 & @5 & @10 
& @5 & @10 & @5 & @10 \\
\hline
CL4S. w/o D & 0.0162          & 0.0268          & 0.0108          & 0.0142          & 0.0444          & 0.0619          & 0.0306          & 0.0363\\
CL4S. $p=0$ & 0.0177          & 0.0292          & 0.0119          & 0.0156          & 0.0451          & 0.0654          & 0.0305          & 0.037\\
CL4S.       & 0.0231          & 0.0369          & 0.0146          & 0.0191          & {\ul 0.0525}    & {\ul 0.0776}    & {\ul 0.0346}    & {\ul 0.0428}    \\
SRMA w/o D  & {\ul 0.0285}    & {\ul 0.0432}    & {\ul 0.0187}    & {\ul 0.0234}    & 0.0504          & 0.0724          & 0.0331          & 0.0402      \\
SRMA w/o M  & 0.0165          & 0.0272          & 0.0104          & 0.0138          & 0.0412          & 0.0590          & 0.0279          & 0.0336     \\
SRMA        & \textbf{0.0299} & \textbf{0.0447} & \textbf{0.0199} & \textbf{0.0246} & \textbf{0.0598} & \textbf{0.0834} & \textbf{0.0407} & \textbf{0.0484} 
\\
\bottomrule
\end{tabular}}
\end{table*}

\subsection{Hyper-parameter Sensitivity}
In this section, we vary the hyper-parameters in neuron masking and layer dropping to draw a detailed investigation of model augmentation. 

\begin{figure}
    \centering
    \includegraphics[width=0.99\linewidth]{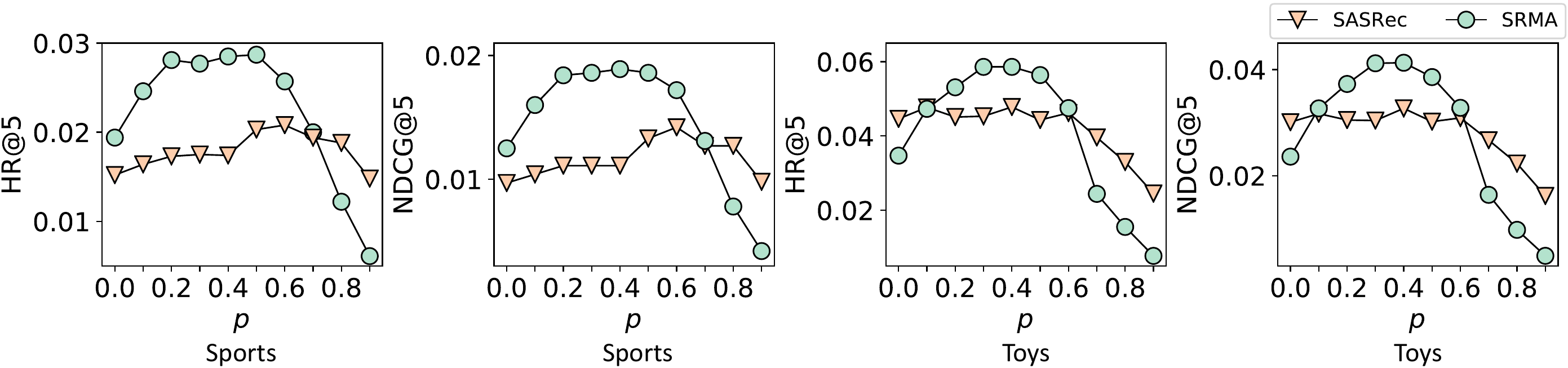}
    \caption{Performance comparison betweeen SASRec and SRMA in HR@5 and NDCG@5 w.r.t different values of neuron masking probability 
    $p$ on Sports and Toys dataset.
    }
    \label{fig:neuron_masking_sensitity}
\end{figure}

\textbf{Effect of Neuron Masking.}
Though all neural layers can apply the neuron masking augmentation, for simplicity, we only endow the FFN layer with the neuron masking augmentation and set the masking probability as $p$ for all FFN layers in the framework.
We fix the settings of layer dropping and the encoder complementing model augmentation and select $p$ from $\{0.0, 0.1, 0.2, \dots, 0.9\}$, where $0.0$ is equivalent to no neuron masking. Also, we compare \modelname with SASRec to justify the effectiveness of the SSL paradigm. The performance curves of HR$@5$ and NDCG$@5$ on the Sports and Toys dataset are demonstrated in Figure~\ref{fig:neuron_masking_sensitity}.  
We can observe that the performance improves first and then drops when increasing $p$ from $0$ to $0.9$. 
The rising of the performance score implies that the neuron masking augmentation is effective in improving the ability of the sequence encoder for a recommendation. 
And the dropping of the performance indicates the intensity of model augmentation should not be overly aggressive, which may lead to less informative contrastive learning. 
As to SASRec, we recognize a higher score of SASRec when $p$ is large, which indicates the optimal model augmentation should be a slightly perturbation rather than a intensive distortion.
Moreover, \modelname consistently outperforms SASRec when $0.1<p<0.6$. Since the only difference is that SASRec has no SSL module, we can thus concludes that the performance gains result from the contrastive SSL step by using the neuron masking.

\begin{figure}
    \centering
    \includegraphics[width=0.99\linewidth]{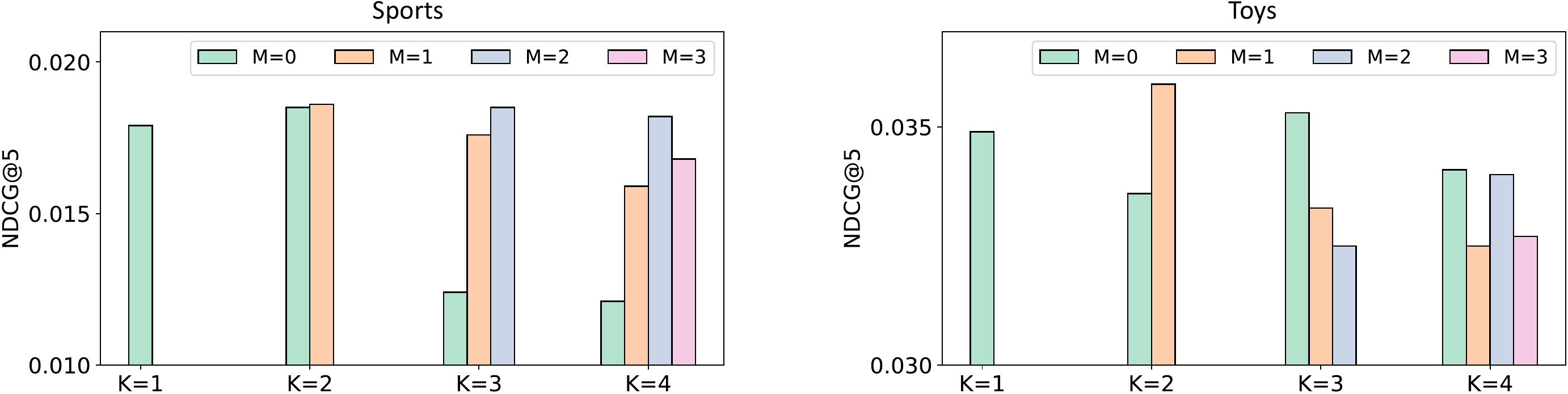}
    \caption{The NDCG@5 performance w.r.t. different $K$ and $M$ for layer dropping augmentation on Sports and Toys dataset.
    }
    \label{fig:layer_dropping}
\end{figure}

\textbf{Effect of Layer Dropping.} The layer dropping model augmentation is controlled by two hyper-parameters, the number of additional FFN layers and the number of layers to drop during training, which is denoted as $K$ and $M$, respectively. Since we can only drop those additional layers, we have $M < K$. We select $K$ from $\{1,2,3,4\}$ while $M$ are searched accordingly. Due to space limitation, we only report the  NDCG$@5$ on the Sports and Toys dataset in Figure~\ref{fig:layer_dropping}. We observe that $K=2, M=1$ achieves the best performance on both datasets, which implies the efficacy of layer dropping. Additionally, we also find that the performance on $K=4$ is consistently worse than $K=2$ on both datasets, which suggests that adding too many layers increases the complexity of the model, which is thus unable to enhance the SSL paradigm.

\subsection{Analyses on Encoder Complementing}
\begin{table*}[htb]
\caption{Performance comparison among SRMA without encoder complementing (w/o Enc.), with Transformer-based (-Trans) and with GRU-based (-GRU) complementary pre-trained encoder. The best score in each column is in bold.}
\vspace{2mm}
\label{tab:ablation-study-ec}
\resizebox{\linewidth}{!}{
\begin{tabular}{c|cc|cc|cc|cc}
\toprule
\multirow{3}{*}{Encoders} & \multicolumn{4}{c|}{Sports} 
& \multicolumn{4}{c}{Toys}\\\cline{2-9}
  
& \multicolumn{2}{c|}{HR} &
\multicolumn{2}{c|}{NDCG} &
\multicolumn{2}{c|}{HR} &
\multicolumn{2}{c}{NDCG} \\
& @5 & @10 & @5 & @10 
& @5 & @10 & @5 & @10 \\
\hline
w/o Enc. & 0.0269 & 0.0401 & 0.0181 & 0.0224 & 0.0567 & 0.0806 & 0.0389 & 0.0466 \\
-Trans & 0.0268 & 0.0408 & 0.0181 & 0.0226 & \textbf{0.0588} & \textbf{0.0811} & \textbf{0.0402} & \textbf{0.0474} \\
-GRU & \textbf{0.0281} & \textbf{0.0411} & \textbf{0.0186} & \textbf{0.0228} & 0.0577 & 0.0811 & 0.0395 & 0.047 \\
\bottomrule
\end{tabular}}
\end{table*}

In this section, we investigate the effects of encoder complementing augmentation for constructing views. Recall that we combine the embedding from the model's encoder and a distinct pre-trained encoder.
For this complementary encoder,
we select from a Transformer-based and a
GRU-based encoder. Since the model's encoder is a 2-layer Transformer, this pre-trained encoder is a 1-layer Transformer to maintain diversity.   
We first pre-train the complementary encoder based on the next-item prediction task.
As such, we empower the pre-trained encoder to characterize the sequential correlations of items. The comparison is conducted on both Sports and Toys datasets, which are
shown in Table~\ref{tab:ablation-study-ec}. The observations are as follows: Firstly, on the Sports dataset, pre-training a GRU encoder as a complement performs the best against the other two, which indicates that injecting distinct encoders for contrastive learning can enhance the SSL signals. 
Secondly, on the Toys dataset, adopting a 1-layer pre-trained Transformer as the complementary encoder yields the best scores on all metrics. 
Besides the effectiveness of encoder complementing, this result also suggests that the complementary encoder may not be overly different from the model's encoder on some datasets,
which otherwise cannot enhance the comprehensive contrastive learning between views.  
Lastly, both Transformer-based and GRU-based pre-trained complementary encoders consistently outperform \modelname without encoder complementing, 
which directly indicates the necessity of encoder complementing as a way of model augmentation.






\section{Conclusion}
This work proposes a novel contrastive self-supervised learning paradigm,  
which simultaneously employs model augmentation and data augmentation to construct views for contrasting. We propose three-level model augmentation methods for this paradigm, which are neuron masking, layer dropping, and encoder complementing.
We adopt this paradigm to the sequential recommendation problem and propose a new model \modelname. 
This model adopts both the random data augmentation of sequences and the corresponding three-level model augmentation for the sequence encoder. 
We conduct comprehensive experiments to verify the effectiveness of \modelname. The overall performance comparison justifies the advantage of contrastive SSL with model augmentation. 
Additionally, detailed investigation regarding the impacts of model augmentation and data augmentation in improving the performance are discussed.
Moreover, ablation studies with respect to three-level model augmentation methods are implemented, which also demonstrate the superiority of the proposed model.
Overall, this work opens up a new direction in constructing views from model augmentation. 
We believe our proposed model augmentation can enhance existing contrastive SSL paradigms which  only have data augmentation.

\newpage

\bibliography{reference}

\begin{thebibliography}{37}
\providecommand{\natexlab}[1]{#1}
\providecommand{\url}[1]{\texttt{#1}}
\expandafter\ifx\csname urlstyle\endcsname\relax
  \providecommand{\doi}[1]{doi: #1}\else
  \providecommand{\doi}{doi: \begingroup \urlstyle{rm}\Url}\fi

\bibitem[Bengio et~al.(2021)Bengio, Lecun, and Hinton]{Bengio2021deep}
Yoshua Bengio, Yann Lecun, and Geoffrey Hinton.
\newblock Deep learning for ai.
\newblock 64\penalty0 (7), 2021.
\newblock ISSN 0001-0782.

\bibitem[Chen et~al.(2020)Chen, Kornblith, Norouzi, and Hinton]{chen2020simple}
Ting Chen, Simon Kornblith, Mohammad Norouzi, and Geoffrey Hinton.
\newblock A simple framework for contrastive learning of visual
  representations.
\newblock In \emph{International conference on machine learning}, pp.\
  1597--1607. PMLR, 2020.

\bibitem[Chen \& He(2021)Chen and He]{chen2021exploring}
Xinlei Chen and Kaiming He.
\newblock Exploring simple siamese representation learning.
\newblock In \emph{Proceedings of the IEEE/CVF Conference on Computer Vision
  and Pattern Recognition}, pp.\  15750--15758, 2021.

\bibitem[Chen et~al.(2018)Chen, Xu, Zhang, Tang, Cao, Qin, and
  Zha]{chen2018sequential}
Xu~Chen, Hongteng Xu, Yongfeng Zhang, Jiaxi Tang, Yixin Cao, Zheng Qin, and
  Hongyuan Zha.
\newblock Sequential recommendation with user memory networks.
\newblock In \emph{WSDM}, pp.\  108--116, 2018.

\bibitem[Dacrema et~al.(2019)Dacrema, Cremonesi, and Jannach]{dacrema2019we}
Maurizio~Ferrari Dacrema, Paolo Cremonesi, and Dietmar Jannach.
\newblock Are we really making much progress? a worrying analysis of recent
  neural recommendation approaches.
\newblock In \emph{Proceedings of the 13th ACM Conference on Recommender
  Systems}, pp.\  101--109, 2019.

\bibitem[Devlin et~al.(2018)Devlin, Chang, Lee, and Toutanova]{devlin2018bert}
Jacob Devlin, Ming-Wei Chang, Kenton Lee, and Kristina Toutanova.
\newblock Bert: Pre-training of deep bidirectional transformers for language
  understanding.
\newblock \emph{arXiv preprint arXiv:1810.04805}, 2018.

\bibitem[Fan et~al.(2021)Fan, Liu, Zhang, Xiong, Zheng, and
  Yu]{fan2021continuous}
Ziwei Fan, Zhiwei Liu, Jiawei Zhang, Yun Xiong, Lei Zheng, and Philip~S. Yu.
\newblock Continuous-time sequential recommendation with temporal graph
  collaborative transformer.
\newblock In \emph{Proceedings of the 30th ACM International Conference on
  Information and Knowledge Management}. ACM, 2021.

\bibitem[Gao et~al.(2021)Gao, Yao, and Chen]{gao2021simcse}
Tianyu Gao, Xingcheng Yao, and Danqi Chen.
\newblock Simcse: Simple contrastive learning of sentence embeddings.
\newblock \emph{arXiv preprint arXiv:2104.08821}, 2021.

\bibitem[He et~al.(2016)He, Zhang, Ren, and Sun]{he2016deep}
Kaiming He, Xiangyu Zhang, Shaoqing Ren, and Jian Sun.
\newblock Deep residual learning for image recognition.
\newblock In \emph{Proceedings of the IEEE conference on computer vision and
  pattern recognition}, pp.\  770--778, 2016.

\bibitem[Hidasi et~al.(2015)Hidasi, Karatzoglou, Baltrunas, and
  Tikk]{hidasi2015session}
Bal{\'a}zs Hidasi, Alexandros Karatzoglou, Linas Baltrunas, and Domonkos Tikk.
\newblock Session-based recommendations with recurrent neural networks.
\newblock \emph{arXiv preprint arXiv:1511.06939}, 2015.

\bibitem[Hu et~al.(2020)Hu, Dong, Wang, Chang, and Sun]{hu2020gpt}
Ziniu Hu, Yuxiao Dong, Kuansan Wang, Kai-Wei Chang, and Yizhou Sun.
\newblock Gpt-gnn: Generative pre-training of graph neural networks.
\newblock In \emph{Proceedings of the 26th ACM SIGKDD International Conference
  on Knowledge Discovery \& Data Mining}, pp.\  1857--1867, 2020.

\bibitem[Hua et~al.(2021)Hua, Wang, Xue, Wang, Ren, and Zhao]{hua2021feature}
Tianyu Hua, Wenxiao Wang, Zihui Xue, Yue Wang, Sucheng Ren, and Hang Zhao.
\newblock On feature decorrelation in self-supervised learning.
\newblock \emph{arXiv preprint arXiv:2105.00470}, 2021.

\bibitem[Kang \& McAuley(2018)Kang and McAuley]{kang2018self}
Wang-Cheng Kang and Julian McAuley.
\newblock Self-attentive sequential recommendation.
\newblock In \emph{ICDM}, pp.\  197--206. IEEE, 2018.

\bibitem[Koch et~al.(2015)Koch, Zemel, Salakhutdinov, et~al.]{koch2015siamese}
Gregory Koch, Richard Zemel, Ruslan Salakhutdinov, et~al.
\newblock Siamese neural networks for one-shot image recognition.
\newblock In \emph{ICML deep learning workshop}, volume~2. Lille, 2015.

\bibitem[Krichene \& Rendle(2020)Krichene and Rendle]{krichene2020sampled}
Walid Krichene and Steffen Rendle.
\newblock On sampled metrics for item recommendation.
\newblock In \emph{SIGKDD}, pp.\  1748--1757, 2020.

\bibitem[Liu et~al.(2020)Liu, Gao, and Ji]{liu2020towards}
Meng Liu, Hongyang Gao, and Shuiwang Ji.
\newblock Towards deeper graph neural networks.
\newblock In \emph{Proceedings of the 26th ACM SIGKDD International Conference
  on Knowledge Discovery \& Data Mining}, pp.\  338--348, 2020.

\bibitem[Liu et~al.(2021{\natexlab{a}})Liu, Zhang, Hou, Mian, Wang, Zhang, and
  Tang]{liu2021self}
Xiao Liu, Fanjin Zhang, Zhenyu Hou, Li~Mian, Zhaoyu Wang, Jing Zhang, and Jie
  Tang.
\newblock Self-supervised learning: Generative or contrastive.
\newblock \emph{IEEE Transactions on Knowledge and Data Engineering},
  2021{\natexlab{a}}.

\bibitem[Liu et~al.(2021{\natexlab{b}})Liu, Chen, Li, Yu, McAuley, and
  Xiong]{liu2021contrastive}
Zhiwei Liu, Yongjun Chen, Jia Li, Philip~S Yu, Julian McAuley, and Caiming
  Xiong.
\newblock Contrastive self-supervised sequential recommendation with robust
  augmentation.
\newblock \emph{arXiv preprint arXiv:2108.06479}, 2021{\natexlab{b}}.

\bibitem[Liu et~al.(2021{\natexlab{c}})Liu, Fan, Wang, and
  Yu]{liu2021augmenting}
Zhiwei Liu, Ziwei Fan, Yu~Wang, and Philip~S. Yu.
\newblock Augmenting sequential recommendation with pseudo-prior items via
  reversely pre-training transformer.
\newblock ACM, 2021{\natexlab{c}}.

\bibitem[McAuley et~al.(2015)McAuley, Targett, Shi, and Van
  Den~Hengel]{mcauley2015image}
Julian McAuley, Christopher Targett, Qinfeng Shi, and Anton Van Den~Hengel.
\newblock Image-based recommendations on styles and substitutes.
\newblock In \emph{SIGIR}, pp.\  43--52, 2015.

\bibitem[Oord et~al.(2016)Oord, Kalchbrenner, Vinyals, Espeholt, Graves, and
  Kavukcuoglu]{oord2016conditional}
Aaron van~den Oord, Nal Kalchbrenner, Oriol Vinyals, Lasse Espeholt, Alex
  Graves, and Koray Kavukcuoglu.
\newblock Conditional image generation with pixelcnn decoders.
\newblock \emph{arXiv preprint arXiv:1606.05328}, 2016.

\bibitem[Oord et~al.(2018)Oord, Li, and Vinyals]{oord2018representation}
Aaron van~den Oord, Yazhe Li, and Oriol Vinyals.
\newblock Representation learning with contrastive predictive coding.
\newblock \emph{arXiv preprint arXiv:1807.03748}, 2018.

\bibitem[Qiu et~al.(2020)Qiu, Chen, Dong, Zhang, Yang, Ding, Wang, and
  Tang]{qiu2020gcc}
Jiezhong Qiu, Qibin Chen, Yuxiao Dong, Jing Zhang, Hongxia Yang, Ming Ding,
  Kuansan Wang, and Jie Tang.
\newblock Gcc: Graph contrastive coding for graph neural network pre-training.
\newblock In \emph{Proceedings of the 26th ACM SIGKDD International Conference
  on Knowledge Discovery \& Data Mining}, pp.\  1150--1160, 2020.

\bibitem[Quadrana et~al.(2017)Quadrana, Karatzoglou, Hidasi, and
  Cremonesi]{quadrana2017personalizing}
Massimo Quadrana, Alexandros Karatzoglou, Bal{\'a}zs Hidasi, and Paolo
  Cremonesi.
\newblock Personalizing session-based recommendations with hierarchical
  recurrent neural networks.
\newblock In \emph{RecSys}, pp.\  130--137, 2017.

\bibitem[Rendle et~al.(2010)Rendle, Freudenthaler, and
  Schmidt-Thieme]{rendle2010factorizing}
Steffen Rendle, Christoph Freudenthaler, and Lars Schmidt-Thieme.
\newblock Factorizing personalized markov chains for next-basket
  recommendation.
\newblock In \emph{WWW}, pp.\  811--820, 2010.

\bibitem[Sun et~al.(2019)Sun, Liu, Wu, Pei, Lin, Ou, and
  Jiang]{sun2019bert4rec}
Fei Sun, Jun Liu, Jian Wu, Changhua Pei, Xiao Lin, Wenwu Ou, and Peng Jiang.
\newblock Bert4rec: Sequential recommendation with bidirectional encoder
  representations from transformer.
\newblock In \emph{CIKM}, pp.\  1441--1450, 2019.

\bibitem[Tang \& Wang(2018)Tang and Wang]{tang2018personalized}
Jiaxi Tang and Ke~Wang.
\newblock Personalized top-n sequential recommendation via convolutional
  sequence embedding.
\newblock In \emph{WSDM}, pp.\  565--573, 2018.

\bibitem[Taylor(1953)]{taylor1953cloze}
Wilson~L Taylor.
\newblock “cloze procedure”: A new tool for measuring readability.
\newblock \emph{Journalism quarterly}, 30\penalty0 (4):\penalty0 415--433,
  1953.

\bibitem[Tian et~al.(2020)Tian, Sun, Poole, Krishnan, Schmid, and
  Isola]{tian2020makes}
Yonglong Tian, Chen Sun, Ben Poole, Dilip Krishnan, Cordelia Schmid, and
  Phillip Isola.
\newblock What makes for good views for contrastive learning?
\newblock \emph{arXiv preprint arXiv:2005.10243}, 2020.

\bibitem[Vaswani et~al.(2017)Vaswani, Shazeer, Parmar, Uszkoreit, Jones, Gomez,
  Kaiser, and Polosukhin]{vaswani2017attention}
Ashish Vaswani, Noam Shazeer, Niki Parmar, Jakob Uszkoreit, Llion Jones,
  Aidan~N Gomez, {\L}ukasz Kaiser, and Illia Polosukhin.
\newblock Attention is all you need.
\newblock In \emph{NIPS}, pp.\  5998--6008, 2017.

\bibitem[Wang et~al.(2019)Wang, He, Wang, Feng, and Chua]{wang2019neural}
Xiang Wang, Xiangnan He, Meng Wang, Fuli Feng, and Tat-Seng Chua.
\newblock Neural graph collaborative filtering.
\newblock In \emph{Proceedings of the 42nd international ACM SIGIR conference
  on Research and development in Information Retrieval}, pp.\  165--174, 2019.

\bibitem[Wu et~al.(2017)Wu, Ahmed, Beutel, Smola, and Jing]{wu2017recurrent}
Chao-Yuan Wu, Amr Ahmed, Alex Beutel, Alexander~J Smola, and How Jing.
\newblock Recurrent recommender networks.
\newblock In \emph{WSDM}, pp.\  495--503, 2017.

\bibitem[Wu et~al.(2020)Wu, Li, Hsieh, and Sharpnack]{ssept20wu}
Liwei Wu, Shuqing Li, Cho-Jui Hsieh, and James Sharpnack.
\newblock Sse-pt: Sequential recommendation via personalized transformer.
\newblock In \emph{RecSys}, pp.\  328–337. ACM, 2020.

\bibitem[Xie et~al.(2020)Xie, Sun, Liu, Gao, Ding, and Cui]{xie2020contrastive}
Xu~Xie, Fei Sun, Zhaoyang Liu, Jinyang Gao, Bolin Ding, and Bin Cui.
\newblock Contrastive pre-training for sequential recommendation.
\newblock \emph{arXiv preprint arXiv:2010.14395}, 2020.

\bibitem[You et~al.(2020)You, Chen, Sui, Chen, Wang, and Shen]{you2020graph}
Yuning You, Tianlong Chen, Yongduo Sui, Ting Chen, Zhangyang Wang, and Yang
  Shen.
\newblock Graph contrastive learning with augmentations.
\newblock \emph{Advances in Neural Information Processing Systems},
  33:\penalty0 5812--5823, 2020.

\bibitem[Zheng et~al.(2019)Zheng, Fan, Lu, Zhang, and Yu]{zheng2019gated}
Lei Zheng, Ziwei Fan, Chun-Ta Lu, Jiawei Zhang, and Philip~S Yu.
\newblock Gated spectral units: Modeling co-evolving patterns for sequential
  recommendation.
\newblock In \emph{SIGIR}, pp.\  1077--1080, 2019.

\bibitem[Zhou et~al.(2020)Zhou, Wang, Zhao, Zhu, Wang, Zhang, Wang, and
  Wen]{zhou2020s3}
Kun Zhou, Hui Wang, Wayne~Xin Zhao, Yutao Zhu, Sirui Wang, Fuzheng Zhang,
  Zhongyuan Wang, and Ji-Rong Wen.
\newblock S3-rec: Self-supervised learning for sequential recommendation with
  mutual information maximization.
\newblock In \emph{Proceedings of the 29th ACM International Conference on
  Information \& Knowledge Management}, pp.\  1893--1902, 2020.

\end{thebibliography}
\bibliographystyle{iclr2022_conference}


\end{document}